\documentclass[sigconf]{acmart} 

\copyrightyear{2021}
\acmYear{2021}
\setcopyright{acmcopyright}
\acmConference[MM '21]{Proceedings of the 29th ACM International Conference on Multimedia}{October 20--24, 2021}{Virtual Event, China}
\acmBooktitle{Proceedings of the 29th ACM International Conference on Multimedia
(MM '21), October 20--24, 2021, Virtual Event, China}
\acmPrice{15.00}
\acmDOI{10.1145/3474085.3475438}
\acmISBN{978-1-4503-8651-7/21/10}
\AtBeginDocument{%
  \providecommand\BibTeX{{%
    \normalfont B\kern-0.5em{\scshape i\kern-0.25em b}\kern-0.8em\TeX}}}
\usepackage{color}

\acmSubmissionID{mfp1510}

\begin{document}
\fancyhead{}

\title{TSA-Net: Tube Self-Attention Network for Action Quality Assessment}

\author{Shunli Wang\textsuperscript{1,3}, Dingkang Yang\textsuperscript{1,2}, Peng Zhai\textsuperscript{1,4}, Chixiao Chen\textsuperscript{1}, Lihua Zhang\textsuperscript{2,1,3,4}}

\authornote{\* indicates corresponding author.}

\affiliation{%
  \institution{Academy for Engineering and Technology, Fudan University\textsuperscript{1}\ \ \ Ji Hua Laboratory, Foshan, China\textsuperscript{2}}
  \city{} \country{}
}
\affiliation{%
  \institution{Engineering Research Center of AI and Robotics, Shanghai, China\textsuperscript{3}}
  \city{} \country{}
}
\affiliation{
  \institution{Jilin Provincial Key Laboratory of Intelligence Science and Engineering, Changchun, China\textsuperscript{4}}
  \city{} \country{}
}
\email{{slwang19, dkyang20, pzhai18, cxchen, lihuazhang}@fudan.edu.cn}

\begin{abstract}
 In recent years, assessing action quality from videos has attracted growing attention in computer vision community and human-computer interaction.
 Most existing approaches usually tackle this problem by directly migrating the model from action recognition tasks, which ignores the intrinsic differences within the feature map such as foreground and background information.
 To address this issue, we propose a Tube Self-Attention Network (TSA-Net) for action quality assessment (AQA). 
 Specifically, we introduce a single object tracker into AQA and propose the Tube Self-Attention Module (TSA), which can efficiently generate rich spatio-temporal contextual information by adopting sparse feature interactions.
 The TSA module is embedded in existing video networks to form TSA-Net.
 Overall, our TSA-Net is with the following merits: 
 1) High computational efficiency,
 2) High flexibility,
 and 3) The state-of-the-art performance. 
 Extensive experiments are conducted on popular action quality assessment datasets including AQA-7 and MTL-AQA.
 Besides, a dataset named Fall Recognition in Figure Skating (FR-FS) is proposed to explore the basic action assessment in the figure skating scene.
 Our TSA-Net achieves the Spearman’s Rank Correlation of 0.8476 and 0.9393 on AQA-7 and MTL-AQA, respectively, which are the new state-of-the-art results. 
 The results on FR-FS also verify the effectiveness of the TSA-Net.
 The code and FR-FS dataset are publicly available at \textit{\url{https://github.com/Shunli-Wang/TSA-Net}}.
\end{abstract}



\maketitle
                    \begin{figure}[t!]
                      \centering
                      \includegraphics[width=\linewidth]{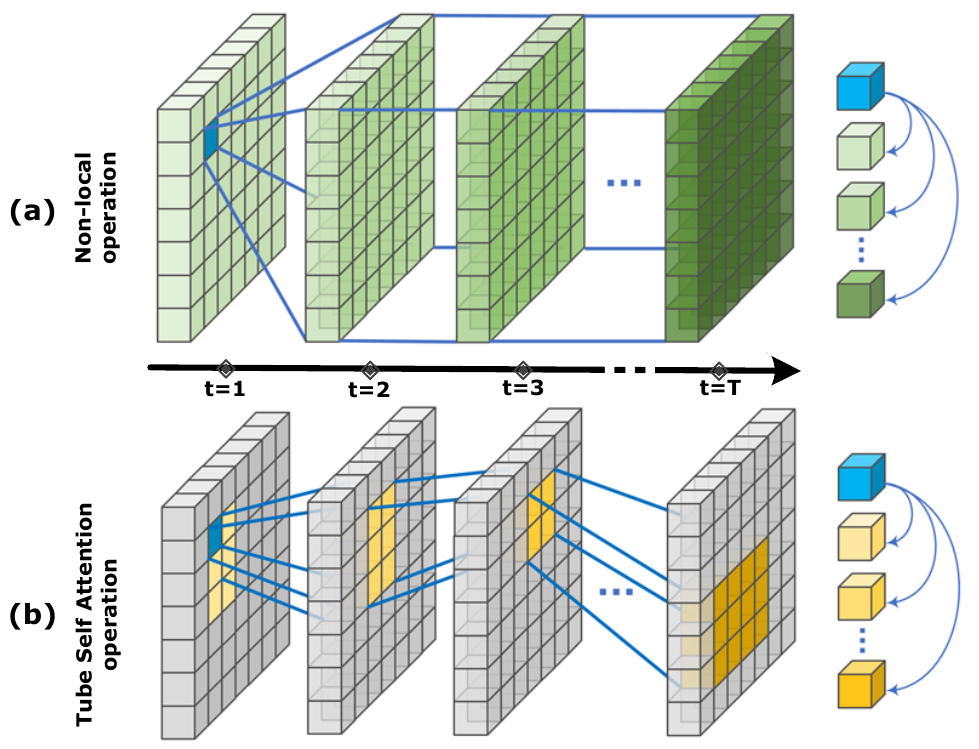}
                      \caption{
                      Diagrams of two attention-based feature aggregation methods. (a) For each position (\textit{e.g.} blue), Non-local module \cite{Nonlocal} performs dense correlation for all features (in green). (b) For each position in the spatio-temporal tube (\textit{e.g.} blue), TSA module only performs sparse correlation for all features located in the tube. 
                      After TSA operation, features in the tube can capture rich contextual information from athletes' series of movements in raw videos. Residual connections are ignored for clear display.
                      }
                      \label{fig1}
                    \end{figure}

\section{Introduction}
 In addition to identifying human action categories in videos, it is also crucial to evaluate the quality of specific actions, which means that the machine needs to understand not only what has been performed but also how well a particular action is performed. 
 Action quality assessment (AQA) aims to evaluate how well a specific action is performed, which has become an emerging and attractive research topic in computer vision community.
 Assessing the quality of actions has great potential value for various real-world applications such as analysis of sports skills \cite{MIT14,PanICCV19,FisV,MUSDL,C3DAVG}, surgical maneuver training \cite{Surgical1,Surgical2,Surgical3} and many others \cite{PairRank18,ProsCons19}.

 In recent years, many methods \cite{PanICCV19,FisV,MUSDL,C3DAVG} directly applied the network of human action recognition (HAR) such as C3D \cite{C3D} and I3D \cite{I3D} to AQA tasks. 
 Although these methods have achieved considerable performance, they still face many challenges, and their performances and efficiency are indeed limited. 
 Firstly, the huge gap between HAR and AQA should be emphasized. Models in HAR require distinguishing subtle differences between different actions, while models in AQA require evaluating a specific action's advantages and disadvantages. Therefore, the performances of existing methods are inherently limited because of the undifferentiated feature extraction of video content, which leads to the pollution of body features. It is not appropriate to apply the framework in HAR directly to AQA without any modification. 
 Secondly, existing methods cannot perform feature aggregation efficiently. The receptive field of convolution operation is limited, resulting in the loss of long-range dependencies. RNN has the inherent property of storing hidden states, which makes it challenging to be paralleled.
 An effective and efficient feature aggregation mechanism is desired in AQA tasks.

 To solve all challenges above, we propose Tube Self-Attention (TSA) module, an efficient feature aggregation strategy based on tube mechanism and self-attention mechanism, shown in Figure \ref{fig1}.
 The basic idea of the TSA module is straightforward and intuitive: considering that AQA models require rich temporal contextual information and do not require irrelevant spatial contextual information, we combine the tube mechanism and self-attention mechanism to aggregate action features sparsely to achieve better performance with minimum computational cost. 
 For example, during a diving competition, the athletes' postures are supposed to raise most attentions, instead of distractors such as the audience and advertisements in the background. 
 The merits of the TSA module are three-fold: (1)\textit{High efficiency}, the tube mechanism makes the network only focus on a subset of the feature map, reducing a large amount of computational complexity compared with Non-local module.
 (2)\textit{Effectiveness}, the self-attention mechanism is adopted in TSA module to aggregate the features in the spatio-temporal tube (ST-Tube), which preserves the contextual information in the time dimension and weakens the influence of redundant spatial information.
 (3)\textit{Flexibility}, consistent with Non-local module, TSA module can be used in a plug-and-play fashion, which can be embedded in any video network with various input sizes.
 
 Based on TSA module, we proposed Tube Self-Attention Network (TSA-Net) for AQA. 
 Existing visual object tracking (VOT) framework is firstly adopted to generate tracking boxes. Then the ST-Tube is obtained through feature selection. The self-attention mechanism is performed in ST-Tube for efficient feature aggregation. 
 Our method is tested on the existing AQA-7 \cite{AQA7} and MTL-AQA \cite{C3DAVG} datasets. Sufficient experimental exploration, including performance analysis and computational cost analysis, is also conducted.
 In addition, a dataset named Fall Recognition in Figure Skating (FR-FS) is proposed to recognize falls in figure skating.
 Experimental results show that our proposed TSA-Net can achieve state-of-the-art results in three datasets. Extensive comparative results verify the efficiency and effectiveness of TSA-Net.

The main contributions of our work are as follows:
\begin{itemize}
    \item 
        We exploit a simple but efficient sparse feature aggregation strategy named Tube Self-Attention (TSA) module to generate representations with rich contextual information for action based on tracking results generated by the VOT tracker.
    \item 
        We propose an effective and efficient action quality assessment framework named TSA-Net based on TSA module, with adding little computational cost compared with Non-local module.
    \item 
        Our approach outperforms state-of-the-arts on the challenging MTL-AQA and AQA-7 datasets and a new proposed dataset named FR-FS. Extensive experiments show that our method has the ability to capture long-range contextual information, which may not be performed by previous methods.
\end{itemize}

                            \begin{figure*}
                              \centering
                              \includegraphics[width=\linewidth]{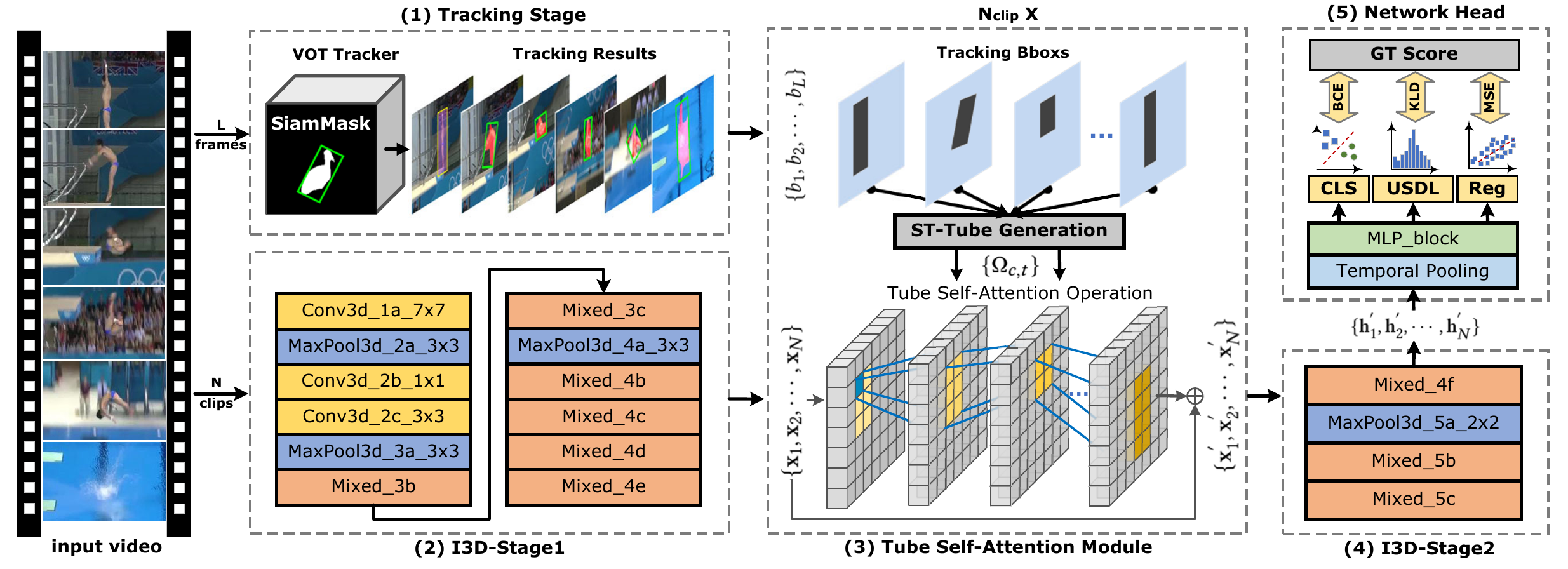}
                              \caption{
                              Overview of the proposed TSA-Net for action quality assessment. 
                              TSA-Net consists of five steps:
                              (1) Tracking. VOT tracker is adopted to generate tracking results \textit{B}.
                              (2) Feature extraction-s1. The input video is divided into \textit{N} clips and the feature extraction is performed by I3D-Stage1 to generate $\mathbf{X}$.
                              (3) Feature aggregation. ST-Tube is generated given $B$ and $\mathbf{X}$, and then the TSA mechanism is used to complete the feature aggregation, results in ${\mathbf{X}}'$.
                              (4) Feature extraction-s2. Aggregated feature ${\mathbf{X}}'$ is passed to I3D-Stage2 to generate ${\mathbf{H}}'$.
                              (5) Network head. The final scores are generated by \textit{MLP\_block}. TSA-Net is trained with different losses according to different tasks.
                              }
                            \label{fig2}
                            \end{figure*}

\section{Related Works}
 \textbf{Action Quality Assessment}. 
 Most of the existing AQA methods focus on two fields: sports video analysis \cite{MIT14,PanICCV19,FisV,MUSDL,C3DAVG} and surgical maneuver assessment  \cite{Surgical1,Surgical2,Surgical3}.
 AQA works focus on sports can be roughly divided into two categories: pose-based methods and non-pose methods. 
 Pose-based methods \cite{PanICCV19,MIT14,BMVC15} take pose estimation results as input to extract features and generate the final scores. 
 Because of the atypical body posture in motion scene, the performance of pose-based methods are suboptimal.

 Non-pose methods exploit DNNs such as C3D and I3D to extract features directly from the raw video and then predict the final score. For example, Self-Attentive LSTM \cite{FisV}, MUSDL \cite{MUSDL}, C3D-AVG-MTL \cite{C3DAVG}, and C3D-LSTM \cite{C3DLSTM} share similar network structures, but their difference lies in the feature extraction and feature aggregation method. Although these methods have achieved significant results, the enormous computational cost of feature extraction and aggregation module limits AQA models' development. 
 Different from the aforementioned AQA methods, our proposed TSA module can perform feature extraction and aggregation efficiently.

\noindent\textbf{Self-Attention Mechanism}.
 Self-attention mechanism \cite{Transformer} was firstly applied on the machine translation task in neural language processing (NLP) as the key part of Transformer. 
 After that, researchers put forward a series of transformer-based models including BERT \cite{BERT}, GPT \cite{GPT}, and GPT-2 \cite{GPT2}. These models tremendously impacted various NLP tasks such as machine translation, question answering system, and text generation. 
 Owing to the excellent performance, some researchers introduce self-attention mechanism into many CV tasks including image classification \cite{Cls1,Cls2,Cls3}, semantic segmentation \cite{Seg1,Seg2,Seg3} and object detection \cite{Det1,Det2,Det3}. 
 Specifically, inspired by Non-local\cite{Nonlocal} module, Huang \textit{et al.} \cite{CCNet} proposed criss-cross attention module and CCNet to avoid dense contextual information in semantic segmentation task. 
 Inspired by these methods, our proposed TSA-Net adopts self-attention mechanism for feature aggregation.

\noindent\textbf{Video Action Recognition}
 Video action recognition is a fundamental task in computer vision. 
 With the rise of deep convolutional neural networks (CNNs) in object recognition and detection, some researchers have designed many deep neural networks for video tasks. 
 Two-stream networks \cite{TwostreamV1, TwostreamV2, I3D} take static images and dynamic optical flow as input and fuse the information of appearance and short-term motions. 
 3D convolutional networks \cite{Ji3D,C3D,LFF2014} utilize 3D kernels to extract features form raw videos directly. 
 In order to meet the needs in real applications, many works \cite{TSM,X3D,R2plus1D} focus on the efficient designing of networks recently.
 The proposed TSA-Net take I3D \cite{I3D} network as the backbone. 
 
\section{Approach}

\subsection{Overview}
The network architecture is given in Figure \ref{fig2}.
 Given an input video with $L$ frames $V=\left \{F_l \right \}_{l=1}^{L}$, SiamMask\cite{SiamMask} is used as the single object tracker to obtain the tracking results $B=\left \{b_l \right \}_{l=1}^{L}$, where $b_l=\{(x_p^l,y_p^l) \}_{p=1}^4$ represents the tracking box of the $l$-th frame. 
 In feature extraction stage, $V$ is firstly divided into $N$ clips where each clip contains M consecutive frames. 
 All clips are further sent into the first stage of Inflated 3D ConvNets (I3D) \cite{I3D}, resulting in $N$ features as $\mathbf{X}=\left \{ \mathbf{x}_n \right \}_{n=1}^N$, $\mathbf{x}_n \in \mathbb{R}^{T\times H\times W\times C}$. 
 Since the temporal length of $\mathbf{x}_n$ is $T$, we have $\mathbf{x}_n =\{\mathbf{x}_{n,t}\}_{t=1}^T$, $\mathbf{x}_{n,t} \in \mathbb{R}^{H\times W\times C}$.

 In feature aggregation stage, the TSA module takes tracking boxes $B$ and video feature $\mathbf{X}$ as input to perform feature aggregation, resulting in video feature ${\mathbf{X}}'= \{ {\mathbf{x}}'_n \}_{n=1}^N$ with rich spatio-temporal contextual information.
 Since the TSA module does not change the size of the input feature map, $\mathbf{x}_n$ and ${\mathbf{x}}'_n$ have the same size, \textit{i.e.}, ${\mathbf{x}}'_n \in \mathbb{R}^{T\times H\times W\times C}$. This property enables TSA modules to be stacked in multiple layers to generate features with richer contextual information.
 The aggregated feature ${\mathbf{X}}'$ is further sent to the second stage of I3D to complete feature extraction, resulting in $\mathbf{H} = \{ \mathbf{h}_n \} _{n=1}^N$. $\mathbf{H}$ is the representation of the whole video or athlete's performance.

 In prediction stage (\textit{i.e.}, network head), average pooling operation is adopted to fuse $\mathbf{H}$ along clip dimension, \textit{i.e.}, $\overline{\mathbf{h}} = \frac{1}{N} \sum_{n=1}^{N} h_n $, $\overline{\mathbf{h}} \in \mathbb{R}^{T\times H\times W\times C}$. 
 $\overline{\mathbf{h}}$ is further fed into the \textit{MLP\_Block} and finally used for the prediction of different tasks according to different datasets.

\subsection{Tube Self-Attention Module} 
The fundamental difference between TSA module and Non-local module is that TSA module can filter the features of participating in self-attention operation in time and space according to the tracking boxes information. 
The TSA mechanism has the ability to ignore noisy background information which will interfere with the final result of action quality assessment.
This operation makes the network pay more attention to the features containing athletes' information and eliminate irrelevant background information interference.

 The tube self-attention mechanism can also be called "local Non-local". 
 The first "local" refers to the ST-Tube, while "Non-local" refers to the response between features calculated by self-attention operation.
 So the TSA module is able to achieve more effective feature aggregation on the premise of saving computing resources.
 TSA module consists of two steps: (1) spatio-temporal tube generation, and (2) tube self-attention operation. 

                        \begin{figure}
                            \centering
                              \includegraphics[width=\linewidth]{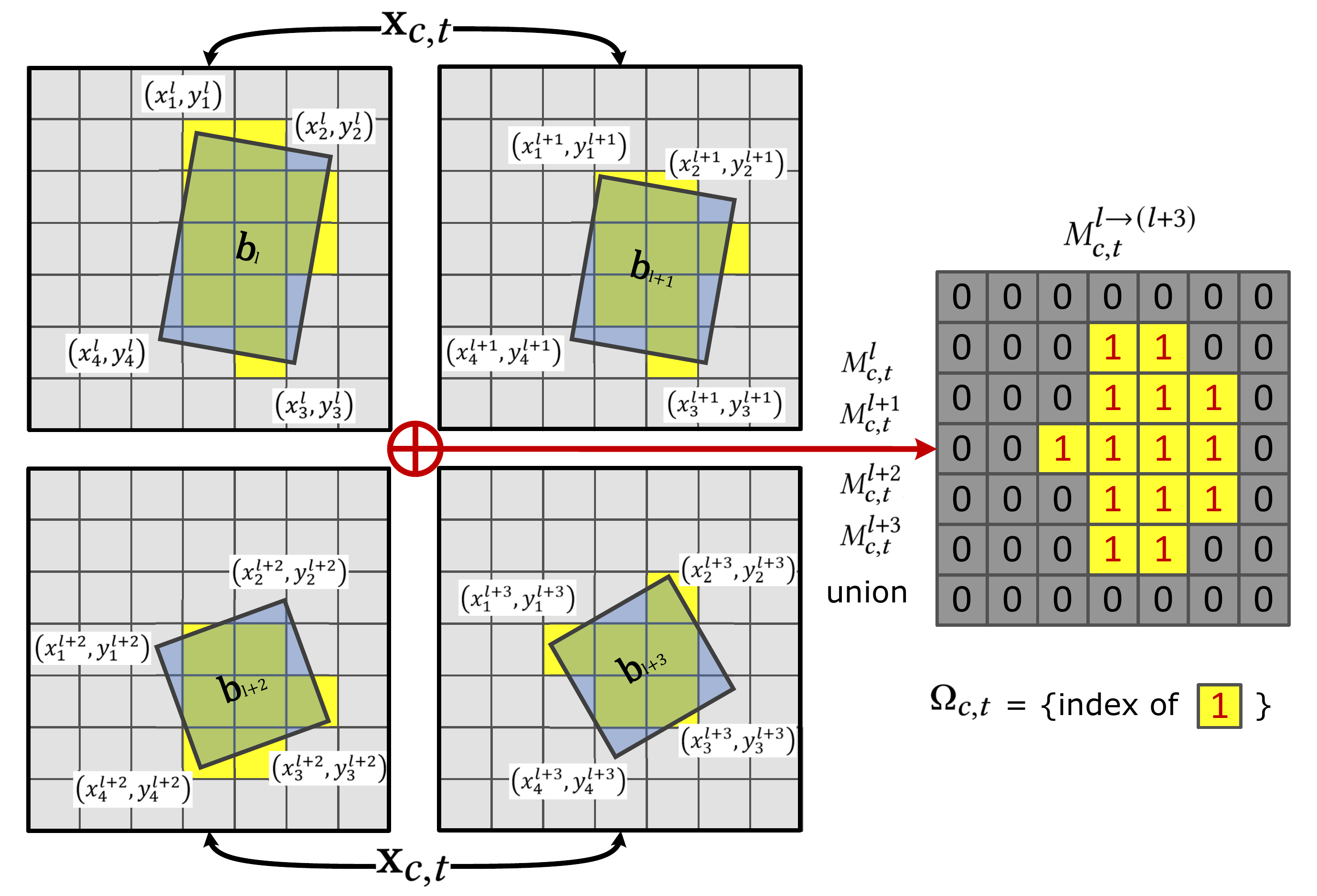}
                              \caption{The generation process of spatio-temporal tube. All boxes $\{b_{l},b_{l+1},b_{l+2},b_{l+3}\}$ are scaled to the same size as the feature map $\mathbf{x}_{c,t}$, and then the separate masks are generated. All masks are aggregated into the final mask $M_{c,t}^{l\rightarrow(l+3)}$ through \textit{Union} operation.}
                              \label{fig3}
                            \end{figure}

\textbf{Step 1: spatio-temporal tube generation.}
Intuitively, after obtaining tracking information $B$ and feature map $\mathbf{X}$ of the whole video, all features in the ST-Tube can be selected directly. 
Unfortunately, owing to the existence of two temporal pooling operations in \textit{I3D-stage1}, the corresponding relationship between tracking boxes and feature maps is not $1:1$ but $many:1$. Besides, all tracking boxes generated by SiamMask are skew, which complicates the generation of ST-Tube. 

 To solve these problems, we propose an alignment method which is shown in Figure \ref{fig3}. 
 Since \textit{I3D-Stage1} contains two temporal pooling operations, the corresponding relationship between bounding boxes and feature map is 4:1, \textit{i.e.}, $\{ b_l, b_{l+1}, b_{l+2}, b_{l+3} \}$ is correspond to $\mathbf{x}_{c,t}$.
 All tracking boxes should be converted into mask first, and then used to generate ST-Tube. 

 We denote the mask of $b_l$ correspond to $\mathbf{x}_{c,t}$ as $M_{c,t}^l \in \{ 0,1 \} ^{H\times W}$.
 The generation process of $M_{c,t}^l$ is as follows:
\begin{equation}
    M_{c,t}^l(i,j) = \left\{\begin{matrix}
    1,  S(b_l,(i,j))\geqslant \tau \\ 
    0,  S(b_l,(i,j)) <  \tau 
    \end{matrix}\right.
\end{equation}
Where $S(\cdot,\cdot)$ function calculates the proportion of the feature grid at $(i,j)$ covered by $b_l$. If the proportion is higher than threshold $\tau $, the feature located at $(i,j)$ will be selected, otherwise it will be discarded. The proportion of each feature grid covered by box ranges from 0 to 1, so we directly took the intermediate value of $\tau = 0.5$ in all experiments of this paper.

Four masks are further assembled into $M_{c,t}^{l\rightarrow (l+3)} \in \{ 0,1 \}^{H \times W}$ through element-wise \textit{OR} operation:
\begin{equation}
    M_{c,t}^{l\rightarrow (l+3)}=Union(M_{c,t}^{l},M_{c,t}^{l+1},M_{c,t}^{l+2},M_{c,t}^{l+3})
\end{equation}
This mask contains all location information of the features participating in self-attention operation.
For the convenience of the following description, $M_{c,t}^{l\rightarrow (l+3)}$ is transformed into the position set of all selected features:
\begin{equation}
    \mathbf{\Omega}_{c,t}=\left \{ (i,j) | M_{c,t}^{l \rightarrow (l+3)}(i,j)=1 \right \}
\end{equation}
Where $\mathbf{\Omega}_{c,t}$ is the basic component of ST-Tube and $\left |\mathbf{\Omega}_{c,t} \right |$ denotes the number of selected features of $\mathbf{x}_{c,t}$.

                \begin{figure}
                  \centering
                  \includegraphics[width=0.68\linewidth]{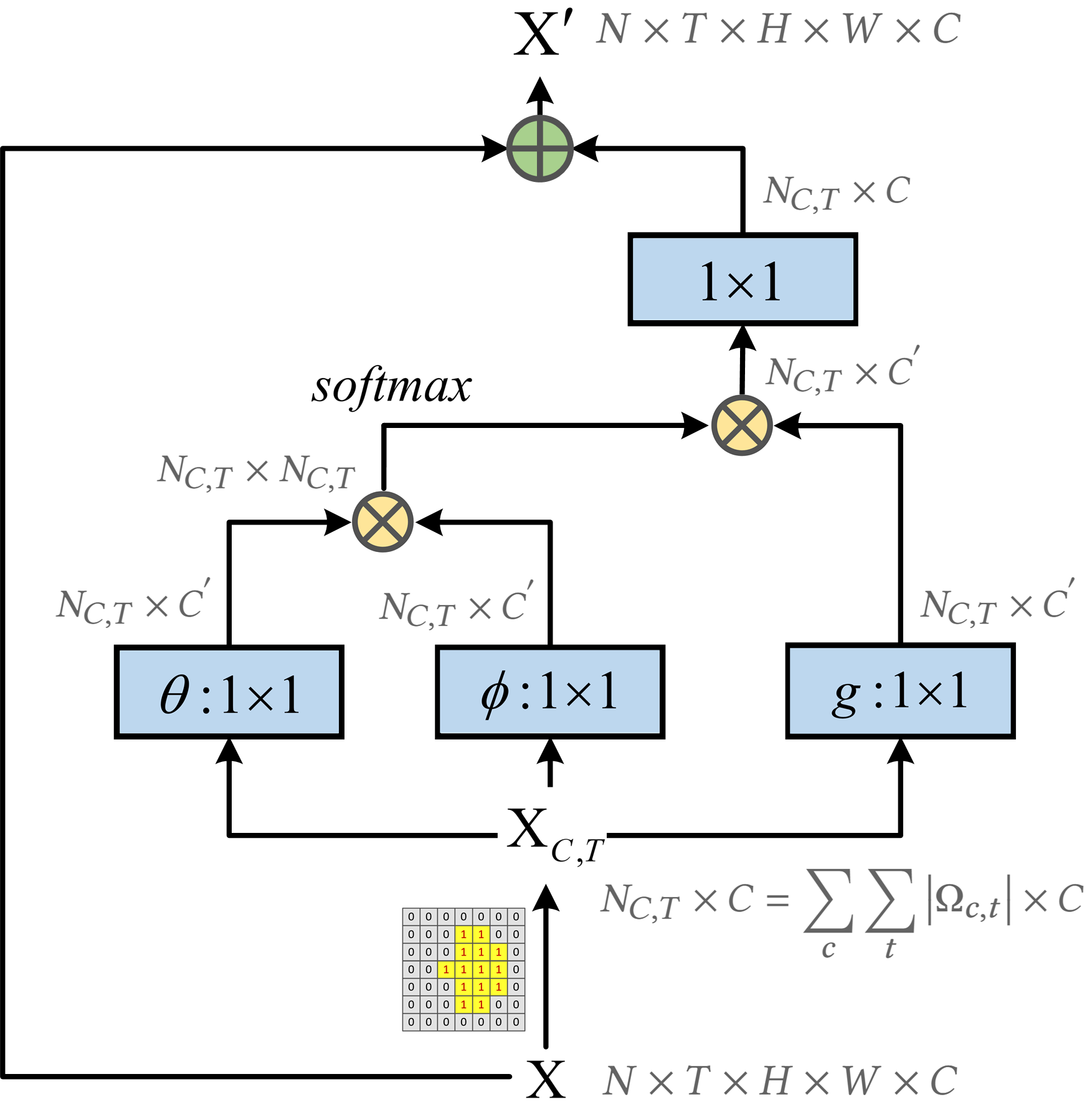}
                  \caption{
                  Calculation process of the TSA module. 
                  "$\oplus$" denotes matrix multiplication, and "$\otimes$" denotes element-wise sum. 
                  Owing to the existence of tube mechanism, only the features inside the ST-Tube can be selected and participate in the calculation of self-attention.}
                  \label{fig4}
                \end{figure}

                \begin{figure*}
                  \centering
                  \includegraphics[width=\linewidth]{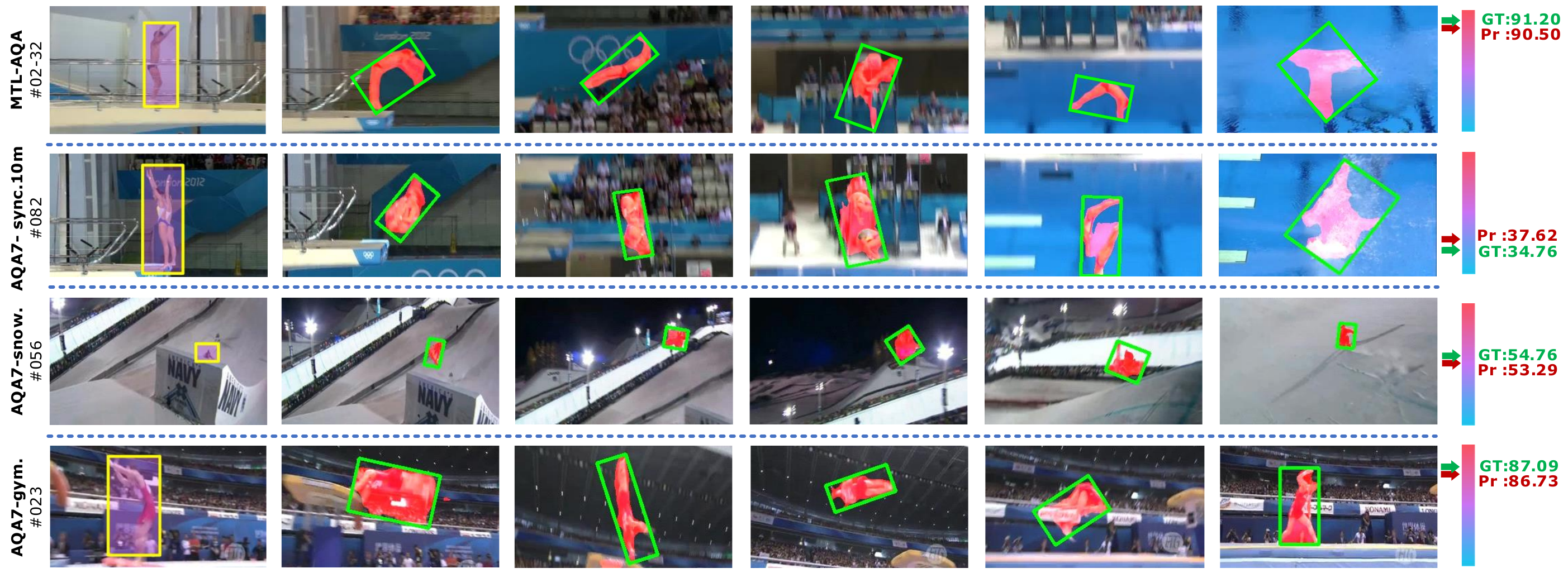}
                  \caption{
                  The tracking results and predicted scores of four cases from four datasets.
                  Four manually annotated initial frames are coloured in yellow, and the subsequent boxes generated by SiamMask are coloured in green.
                  The predicted scores of TSA-Net and GT scores are shown on the right.
                  More visualization cases can be found in supplementary materials.}
                  \label{fig5}
                \end{figure*}

\textbf{Step 2: tube self-attention operation} %
After obtaining $\mathbf{X}$ and $\mathbf{\Omega}_{c,t}$, the self-attention mechanism is performed to aggregate all features located in ST-Tube, as shown in Figure \ref{fig4}.
The formation of the TSA mechanism adopted in this paper is consistent with \cite{Nonlocal}:

\begin{equation}
    \mathbf{y}_p=\frac{1}{C(\mathbf{x})}\sum_{\forall c}\sum_{\forall t}\sum_{\forall (i,j)\in \mathbf{\Omega }_{c,t}}f\left (\mathbf{x}_p,\mathbf{x}_{c,t}(i,j))g(\mathbf{x}_{c,t}(i,j) \right )
\end{equation}
Where $p$ denotes the index of an output position whose response is to be computed. $(c,t,i,j)$ is the input index that enumerates all positions in ST-Tube. Output feature map $\mathbf{y}$ and input feature map $\mathbf{x}$ have the same size. $f(\cdot,\cdot)$ denotes the pairwise function, and $g(\cdot)$ denotes the unary function. The response is normalized by $C(\mathbf{x})=\sum_c \sum_t \left | \mathbf{\Omega }_{c,t}\right |$.

To reduce the computational complexity, the dot product similarity function is adopted:
\begin{equation}
    f\left ( \mathbf{x}_p, \mathbf{x}_{c,t}(i,j) \right ) = \theta (\mathbf{x}_p)^T\phi (\mathbf{x}_{c,t}(i,j))
\end{equation}
Where both $\theta(\cdot)$ and $\phi(\cdot)$ are channel reduction transformations.

Finally, the residual link is added to obtain the final $\mathbf{X}'$:
\begin{equation}
    \mathbf{x}'_p = W_z\mathbf{y}_p + \mathbf{x}_p
\end{equation}
Where $W_z\mathbf{y}_p$ denotes an embedding of $\mathbf{y}_p$. 
Note that $\mathbf{x}'_p$ has the same size with $\mathbf{x}_p$, so TSA module can be inserted into any position in deep convolutional neural networks.
For the trade-off between computational cost and performance, all TSA modules are placed after \textit{Mixed\_4e}. Thus, $T=4$ and $H=W=14$.

Compared with the Non-local operation, the TSA module greatly reduces the computational complexity in time and space from
\begin{equation}
    O\left ( (N\times T\times H\times W) \times (N\times T\times H\times W) \right )
\end{equation}
to 
\begin{equation}
    O\left ( \left ( \sum_c \sum_t \left | \mathbf{\Omega }_{c,t}\right | \right )\times \left ( \sum_c \sum_t \left | \mathbf{\Omega }_{c,t}\right | \right ) \right )
\end{equation}
 Note that the computational cost of TSA can only be measured after forwarding propagation because $\mathbf{\Omega }_{c,t}$ is generated from $B$.

\subsection{Network Head and Training}
To verify the effectiveness of the TSA module, we extend the network head to support multiple tasks, including classification, regression, and score distribution prediction. 
All tasks can be achieved by changing the output size of \textit{MLP\_block} and the definition of the loss function.
The implementation details of these three tasks are as follows:

\textbf{Classification}. When dealing with classification tasks, the output dimension of \textit{MLP\_block} is determined by the number of categories. Binary Cross-Entropy loss (BCELoss) is adopted.

\textbf{Regression}. When dealing with regression tasks, the output dimension of \textit{MLP\_block} is set to 1. Mean Square Error loss (MSELoss) is adopted.

                    \begin{table*}
                      \caption{Comparison with state-of-the-arts on AQA-7 Dataset.}
                      \label{tab1}
                      \begin{tabular}{c|cccccc|c}
                        \toprule
                        Method & Diving & Gym Vault & Skiing & Snowboard & Sync. 3m & Sync. 10m  & Avg. Corr.\\
                        \midrule
                        Pose+DCT \cite{MIT14}    & 0.5300 & - & - & - & - & - & - \\
                        ST-GCN \cite{STGCN18}      & 0.3286 & 0.577 & 0.1681 & 0.1234 & 0.6600 & 0.6483 & 0.4433 \\
                        C3D-LSTM \cite{C3DLSTM}    & 0.6047 & 0.5636 & 0.4593 & 0.5029 & 0.7912 & 0.6927 & 0.6165 \\
                        C3D-SVR \cite{C3DLSTM}     & 0.7902 & 0.6824 & 0.5209 & 0.4006 & 0.5937 & 0.9120 & 0.6937 \\
                        JRG \cite{PanICCV19}         & 0.7630 & 0.7358 & 0.6006 & 0.5405 & 0.9013 & 0.9254 & 0.7849 \\
                        USDL \cite{MUSDL}        & 0.8099 & 0.757 & 0.6538 & \textbf{0.7109} & 0.9166 & 0.8878 & 0.8102 \\
                        \midrule
                        NL-Net      & 0.8296 & 0.7938 & \textbf{0.6698} & 0.6856 & 0.9459 & 0.9294 & 0.8418 \\
                        TSA-Net (Ours) & \textbf{0.8379} & \textbf{0.8004} & 0.6657 & 0.6962 & \textbf{0.9493} & \textbf{0.9334} & \textbf{0.8476} \\
                      \bottomrule
                    \end{tabular}
                    \end{table*}

\textbf{Score distribution prediction}. 
Tang \textit{et al.} \cite{MUSDL} proposed an uncertainty-aware score distribution learning (USDL) approach and its multi-path version MUSDL for AQA tasks. 
Although experiment results in \cite{MUSDL} proved the superiority of MUSDL compared with USDL, a multi-path strategy will lead to a significant increase in computational cost. 
However, the TSA module can generate features with rich contextual information by adopting a self-attention mechanism in ST-Tube with less computational complexity.

To verify the effectiveness of the TSA module, we embed the TSA module into the USDL model.
The loss function is defined as Kullback-Leibler (KL) divergence of predicted score distribution and ground-truth (GT) score distribution:
\begin{equation}
   KL\left \{ p_c \parallel s_{pre} \right \}=\sum_{i=1}^{m}p(c_i)log\frac{p(c_i)}{s_{pre}(c_i)} 
\end{equation}
Where $s_{pre}$ is generated by \textit{MLP\_block}, and $p_c$ is generated by GT score.
 Note that for dataset with difficulty degree (DD), $s=DD \times s_{pre}$ is used as the final predicted score.

                    \begin{table*}
                      \caption{Study on different settings of the number of TSA module.}
                      \label{tab:AQA7-2}
                      \begin{tabular}{c|cccccc|c}
                        \toprule
                        Method & Diving & Gym Vault & Skiing & Snowboard & Sync. 3m & Sync. 10m  & Avg. Corr.\\
                        \midrule
                        TSA-Net & 0.8379 & 0.8004 & 0.6657 & 0.6962 & \textbf{0.9493} & 0.9334 & 0.8476 \\
                        TSAx2-Net & 0.8380 & 0.7815 & \textbf{0.6849} & \textbf{0.7254} & 0.9483 & \textbf{0.9423} & \textbf{0.8526} \\
                        TSAx3-Net & \textbf{0.8520} & \textbf{0.8014} & 0.6437 & 0.6619 & 0.9331 & 0.9249 & 0.8352 \\
                      \bottomrule
                    \end{tabular}
                    \end{table*}
                    
                    \begin{table}
                      \caption{Comparisons of computational complexity and performance on AQA-7. GFLOPs is adopted to measure the computational cost.}
                      \label{tab3}
                      \begin{tabular}{c|cc|cc}
                        \toprule
                        Method & NL-Net & TSA-Net & Comp. Dec. & Corr. Imp.\\
                        \midrule
                        Diving      & 2.2G & 0.864G & -60.72\% & $\uparrow$0.0083 \\
                        Gym Vault   & 2.2G & 0.849G & -61.43\% & $\uparrow$0.0066 \\
                        Skiing      & 2.2G & 0.283G & -87.13\% & $\downarrow$0.0041 \\
                        Snowboard   & 2.2G & 0.265G & -87.97\% & $\uparrow$0.0106 \\
                        Sync. 3m    & 2.2G & 0.952G & -56.74\% & $\uparrow$0.0034 \\
                        Sync. 10m   & 2.2G & 0.919G & -58.24\% & $\uparrow$0.0040 \\
                        \midrule
                        Average     & 2.2G & 0.689G & -68.70\% & $\uparrow$0.0058\\
                      \bottomrule
                    \end{tabular}
                    \end{table}

\section{Experiments} \label{Experiments}
We carry out comprehensive experiments on AQA-7 \cite{AQA7}, MTL-AQA \cite{C3DAVG}, and FR-FS datasets to evaluate the proposed method. 
Experimental results demonstrate that TSA-Net achieves state-of-the-art performance on these datasets. 
In the following subsections, we first introduce two public datasets and a new dataset named Fall Detection in Figure Skating (FD-FS) proposed by us. 
After that, a series of experiments and computational complexity analysis are performed on AQA-7 and MTL-AQA datasets. 
Finally, the detection results on FD-FS are reported, and the network prediction results are analyzed visually and qualitatively.

\subsection{Datasets and Evaluation Metrics}

\textbf{AQA-7} \cite{AQA7}. 
 The AQA-7 dataset comprising samples from seven actions.
 It contains 1189 videos, in which 803 videos are used for training and 303 videos used for testing. 
 To ensure the comparability with other models, we delete \textit{trampoline} category because of its long time.

\noindent\textbf{MTL-AQA} \cite{C3DAVG}. 
 The MTL-AQA dataset is currently the largest dataset for AQA tasks. There are 1412 diving samples collected from 16 different events in MTL-AQA.
 Furthermore, MTL-AQA provides detailed scoring of each referee, diving difficulty degree, and live commentary.
 We followed the evaluation protocol suggested in \cite{C3DAVG}, so that there are 1059 samples used for training and 353 used for testing.

\noindent\textbf{FR-FS (Fall Recognition in Figure Skating)}.
 Although some methods have been proposed \cite{MIT14,FisV} to evaluate figure skating skills, they are only based on long-term videos which last nearly 3 minutes.
 These coarse-grained methods will lead to the inundation of detailed information in a long time scale. 
 However, these details are crucial and indispensable for AQA tasks.
 To address this issue, we propose a dataset named FR-FS to recognize falls in figure skating sports. 
 We plan to start from the most basic fault recognition and gradually build a more delicate granularity figure skating AQA system. 

 The FR-FS dataset contains 417 videos collected from FIV \cite{FisV} and \textit{Pingchang 2018 Winter Olympic Games}. FR-FS contains the critical movements of the athlete's take-off, rotation, and landing. Among them, 276 are smooth landing videos, and 141 are fall videos. 
 To test the generalization performance of our proposed model, we randomly select 50\% of the videos from the fall and landing videos as the training set and the testing set. 

\noindent\textbf{Evaluation Protocols}.
 Spearman's rank correlation is adopted as the performance metric to measure the divergence between the GT score and the predicted score. 
 The Spearman’s rank correlation is defined as follows:
\begin{equation}
  \rho =\frac{\sum (p_i-\bar{p})(q_i-\bar{q})}{\sqrt{\sum(p_i-\bar{p})^2\sum(q_i-\bar{q})^2}}
\end{equation}
 Where $p$ and $q$ represent the ranking of GT and predicted score series, respectively. Fisher’s z-value \cite{C3DLSTM} is used to measure the average performance across multiple actions.

                    \begin{table}
                      \caption{Comparison with state-of-the-arts on MTL-AQA.}
                      \label{tab4}
                      \begin{tabular}{cc}
                        \toprule
                        Method & Avg. Corr. \\
                        \midrule
                        Pose+DCT \cite{MIT14}        & 0.2682\\
                        C3D-SVR \cite{C3DLSTM}         & 0.7716\\
                        C3D-LSTM \cite{C3DLSTM}        & 0.8489\\
                        C3D-AVG-STL \cite{C3DAVG}     & 0.8960\\
                        C3D-AVG-MTL \cite{C3DAVG}     & 0.9044\\
                        MUSDL \cite{MUSDL}           & 0.9273\\
                        \midrule
                        NL-Net          & \textbf{0.9422}\\
                        TSA-Net         & 0.9393\\
                      \bottomrule
                    \end{tabular}
                    \end{table}

\subsection{Implementation Details}
 Our proposed methods were built on the Pytorch toolbox \cite{Pytorch} and implemented on a system with the Intel (R) Xeon (R) CPU E5-2698 V4 @ 2.20GHz. 
 All models are trained on a single NVIDIA Tesla V100 GPU.
 Faster-RCNN\cite{FasterRCNN} pretrained on MS-COCO\cite{MSCOCO} is adopted to detect the athletes in all initial frames.
 All videos are normalized to $L=103$ frames.
 For all experiments, the I3D\cite{I3D} pretrained on Kinetics \cite{Kinetics} is utilized as the feature extractor. 
 All videos are select from high-quality sports broadcast videos, and the athletes' movements are apparent. Therefore, we argue that the performance of TSA-Net is not sensitive to the choice of the tracker. SiamMask\cite{SiamMask} was chosen only for high-speed and tight boxes.
 Each training mini-batch contains 4 samples.
 Adam \cite{Adam} optimizer was adopted for network optimization with initial learning rate 1e-4, momentum 0.9, and weight decay 1e-5.

                        \begin{figure}
                          \centering
                          \includegraphics[width=\linewidth]{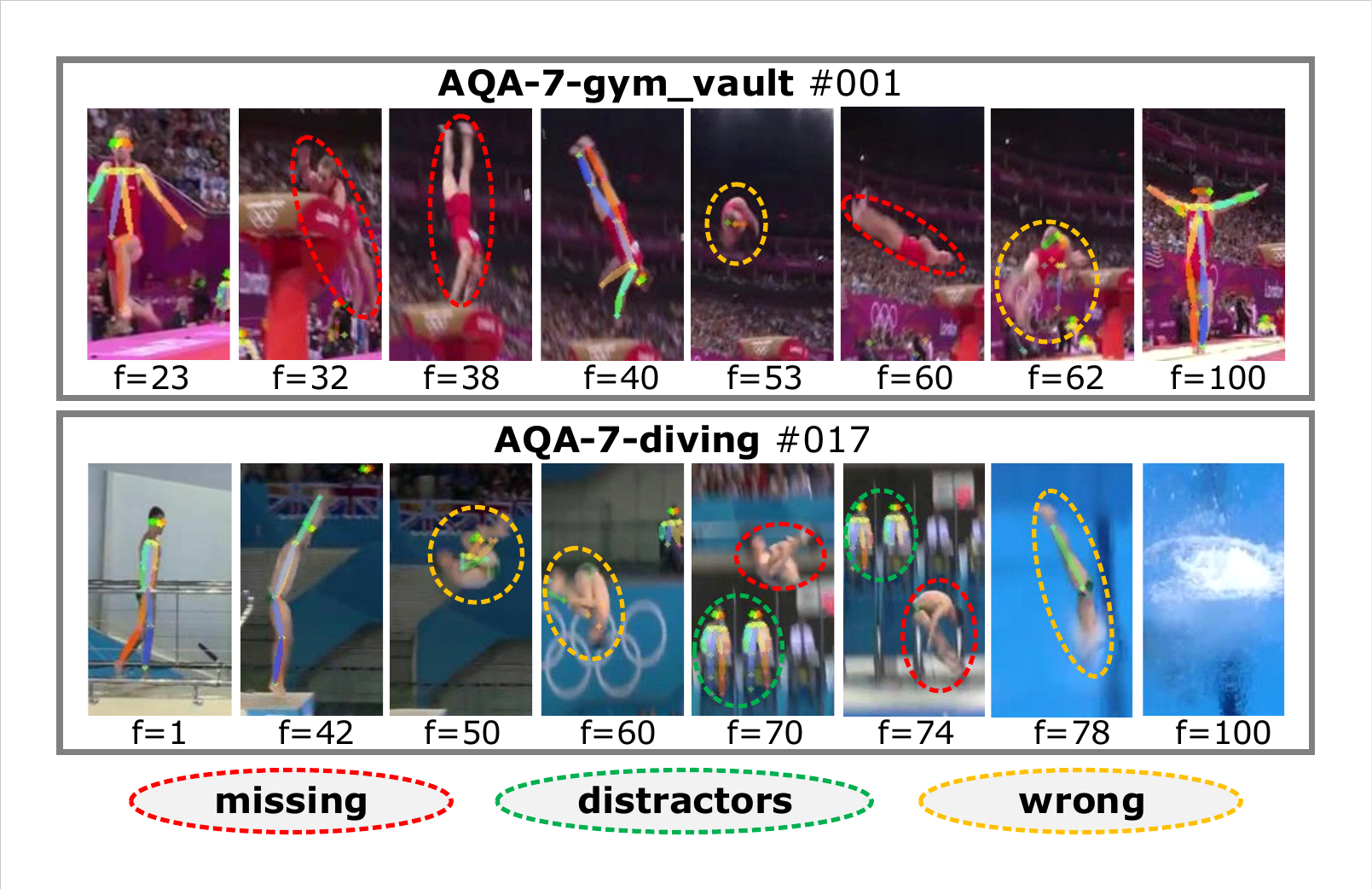}
                          \caption{Alphapose \cite{Alphapose} is selected as the pose estimator. The estimation results of two sports videos are visualized.}
                          \label{Pose-fail}
                        \end{figure}

 Considering the complexity differences between datasets, we adopt different experimental settings.
 In AQA-7 and MTL-AQA datasets, all videos are divided into 10 clips consistent with \cite{MUSDL}.
 Random horizontal flipping and timing offset are performed on videos in training phase. Training epoch is set to 100. All video score normalization are consistent with USDL \cite{MUSDL}.
 In FR-FS dataset, all videos are divided into 7 segments to prevent overfitting. Training epoch is set to 20. 

\subsection{Results on AQA-7 Dataset}

 The TSA module and the Non-local module are embedded after \textit{Mixed\_4e} of I3D to create TSA-Net and NL-Net.
 Experimental results in Table \ref{tab1} show that TSA-Net achieves 0.8476 on Avg. Corr. , which is higher than 0.8102 of USDL.
 TSA-Net outperforms USDL in all categories except the \textit{snowboard}. This is mainly caused by the size issue: the small size of the target leads to the small size of the ST-Tube, resulting in invalid feature enhancement (AQA-7 \textit{snow. \#056} in Figure \ref{fig5}).
 Note that the TSA module is used in a plug-and-play fashion, comparative experiments in Table \ref{tab1} can also be regarded as ablation studies. Therefore, we didn't set up a separate part of ablation in this paper.

\noindent\textbf{The effect of different number of TSA module}. 
 Inspired by the multi-layer attention mechanism in Transformer \cite{Transformer}, we stack multiple TSA modules and test these variants on AQA-7.
 Experimental results in Table \ref{tab:AQA7-2} show that TSA-Net achieves the best performance when $N_{stack} = 2$. Benefit from the feature aggregation operations conducted by two subsequent TSA modules, the network can capture richer contextual features compared with USDL.
 When $N_{stack} = 3$, the performance of the model becomes worse, which may be caused by overfitting.

\noindent\textbf{Computational cost analysis}. 
 Computational cost comparison results are shown in Table \ref{tab3}. 
 Note that only the calculation of TSA module or Non-local module is counted, not the whole network.
 Compared with NL-Net, TSA-Net can reduce the computation by 68.7\% on average and bring 0.0058 AVG. Corr. improvement. 
 This is attributed to the tube mechanism adopted in TSA module, which can avoid dense attention calculation and improve performance simultaneously.
 Among all categories in AQA-7, the TSA module saves up to 87\% of the computational complexity on \textit{skiing} and \textit{snowboard}. 
 Such a large reduction is caused by the small size of the ST-Tube.
 However, the small ST-Tube will hinder the network from completing effective feature aggregation and ultimately affect the final performance.
 This conclusion is consistent with the analysis of the results in Table \ref{tab1}.

                \begin{figure*}
                  \centering
                  \includegraphics[width=\linewidth]{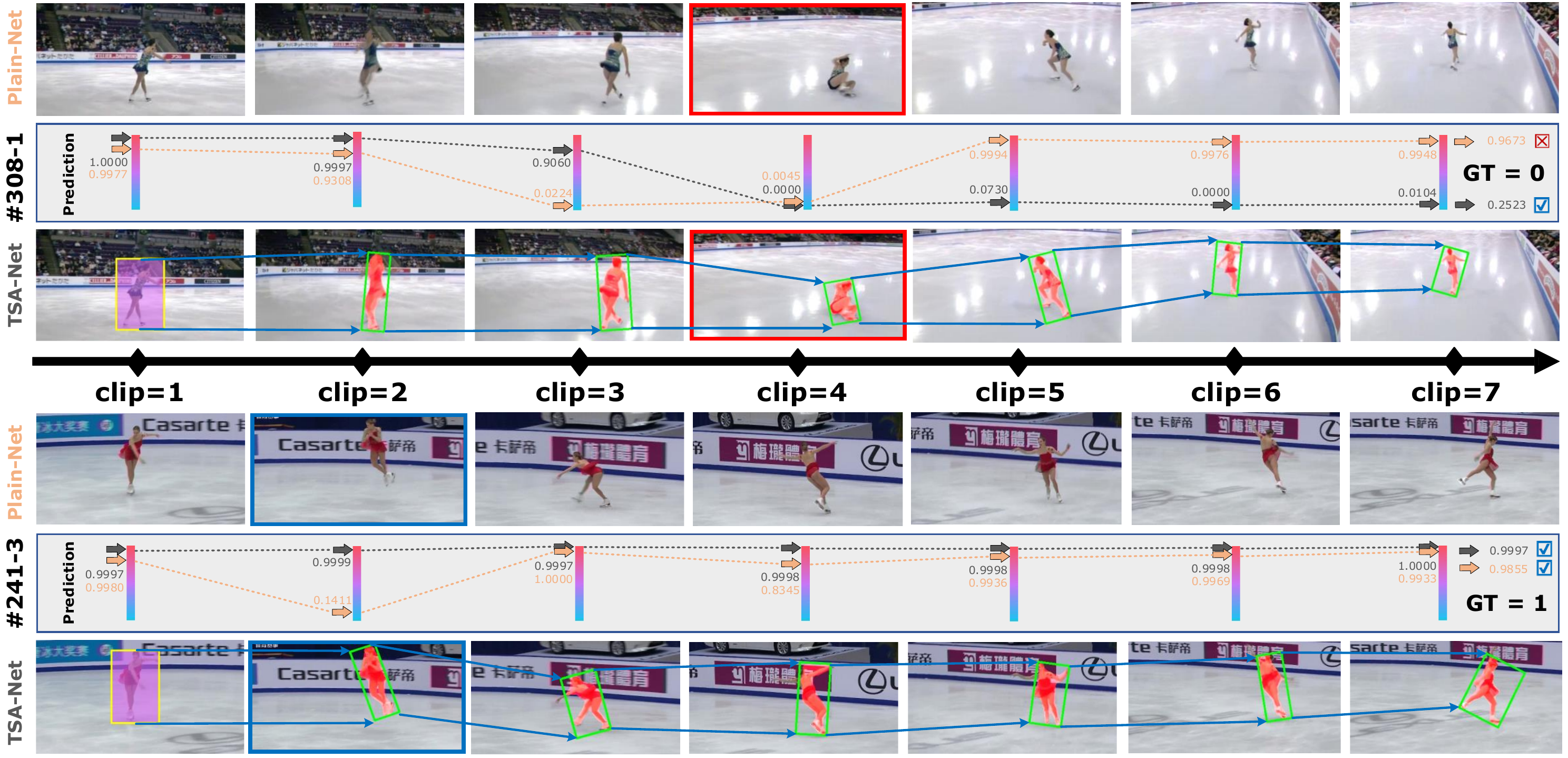}
                  \caption{Case study with qualitative results on FR-FS. The failure case \#308-1 is above the timeline, while the successful case \#241-3 is below the timeline.}
                  \label{fig7}
                \end{figure*}

\subsection{Results on MTL-AQA Dataset}
 As shown in Table \ref{tab4}, the TSA-Net and NL-Net is compared with existing methods. Regression network head and MSELoss are adopted in two networks.
 Experimental results show that both TSA-Net and NL-Net can achieve state-of-the-art performance and the NL-Net is better. 
 The performance fluctuation of TSA-Net is mainly caused by different data distribution between two datasets. 
 Videos in MTL-AQA have higher resolution (640x360 to 320x240) and broader field of view, which leads to smaller ST-Tubes in TSA-Net and affects the performance. 
 It should be emphasized that this impact is feeble. 
 TSA-Net saves half of the computational cost and achieves almost the same performance as NL-Net. 
 This proves the effectiveness and efficiency of the TSA-Net, which is not contradictory to the final conclusion.

\noindent\textbf{Studys on the the stack number of TSA modules and computational cost}.
 As shown in Tabel \ref{tab5}, three parallel experiments are conducted with only the number of TSA module changed just as the experiments on AQA-7. 
 If NL-Net is excluded, the best \textit{Sp. Corr.} is achieved when $N_{clip}=2$ (\textit{i.e.}, TSAx2-Net), while TSA-Net with only one TSA module achieves minimum MSE and computational cost simultaneously. 
 This phenomenon is mainly caused by low computational complexity of TSA-Net. 
 The sparse feature interaction characteristics of TSA module achieve more efficient feature enhancement and have the ability to avoid overfitting.
 Although the performance of TSA-Net can be improved by increasing the number of TSA modules, it will increase computational cost.
 To achieve the balance between computational cost and performance, we only take $N_{stack} = 1$ in all subsequent experiments. 

                    \begin{table}
                      \caption{Comparisons of computational complexity and performance between NL-Net and the variants of TSA-Net on MTL-AQA.}
                      \label{tab5}
                      \begin{tabular}{cccc}
                        \toprule
                        Method & Sp. Corr.$\uparrow$ & MSE$\downarrow$ & FLOPs$\downarrow$\\
                        \midrule
                        NL-Net      & \textbf{0.9422} & 47.83 & 2.2G \\
                        TSA-Net     & 0.9393 & \textbf{37.90} & \textbf{1.012}G\\
                        TSAx2-Net   & 0.9412 & 46.51 & 2.025G\\
                        TSAx3-Net   & 0.9403 & 47.77 & 3.037G\\
                        
                      \bottomrule
                    \end{tabular}
                    \end{table}

\subsection{Results on FR-FS Dataset}

                    \begin{table}
                      \caption{Recognition accuracy on FR-FS.}
                      \label{tab6}
                      \begin{tabular}{cc}
                        \toprule
                        Method & Acc.\\
                        \midrule
                        Plain-Net   & 94.23 \\
                        TSA-Net & \textbf{98.56} \\
                      \bottomrule
                    \end{tabular}
                    \end{table}

 In FR-FS dataset, we focus on the performance improvement that the TSA module can achieve. 
 Therefore, Plain-Net and TSA-Net are implemented, respectively. The former does not adopt any feature enhancement mechanism, while the latter is equipped with a TSA module. 
 As shown in Table \ref{tab6}, TSA-Net outperforms Plain-Net by 4.33\%, which proves the effectiveness of TSA module.

\noindent\textbf{Visualization of Temporal Evolution.} 
 A case study is also conducted to further explore the performance of TSA-Net. 
 Two representative videos are selected and the prediction results of each video clip are visualized in Figure \ref{fig7}, .
 All clip scores are obtained by deleting the temporal pooling operation in Plain-Net and TSA-Net. 
 In the failure case \#308-1, both Plain-Net and TSA-Net can detect that the athlete falls in the fourth chip which highlighted in red, but only TSA-Net gets the correct result in the end (0.9673 for Plain-Net and 0.2523 for TSA-Net). 
 The TSA mechanism forces the features in ST-Tube interact with each other in the way of self-attention, which makes TSA-Net regard the standing up and adjusting actions after falling as errors in clip 5 to 7.

 It seems that TSA-Net is too strict in fall recognition, but the analysis in the successful case \#241-3 has overturned this view.
 Two models get similar results, except for the second clip (colored in blue), which contains the take-off and rotation phase. 
 Plain-Net has great uncertainty for the stationarity of take-off phase, while TSA-Net can get high confidence results. 
 Based on visual analysis and quantitative analysis, it can be concluded that the TSA module is able to perform feature aggregation effectively and obtain more reasonable and stable prediction results.
 
\balance
\subsection{Analysis and Visualization}

\textbf{Reasons for choosing tracking boxes over pose estimation.}
 In sports scenes, high-speed movements of the human body will lead to a series of challenges such as motion blur and self-occlusion, which eventually result in failure cases in pose estimation. 
 Results in Figure \ref{Pose-fail} show that Alphapose \cite{Alphapose} cannot handle these situations properly. 
 The missing of human posture and background audience posture interference  will seriously affect the evaluation results.
 Previous studies in FineGym \cite{FineGym} have come to the same results as ours. 
 Based on these observations, we conclude that methods based on pose estimation are not suitable for AQA in sports scenes. Missing boxes and wrong poses will significantly limit the performance of the AQA model.
 Therefore, we naturally introduce the VOT tracker into AQA tasks.
 The proposed TSA-Net achieves significant improvement in AQA-7 and MTL-AQA compared to pose-based methods such as Pose+DCT \cite{MIT14} and ST-GCN \cite{STGCN18} as shown in Table \ref{tab1} and \ref{tab4}. 
 These comparisons show that the TSA mechanism is superior to the posture-based mechanism in capturing key dynamic characteristics of human motion.

\noindent\textbf{Visualization on MTL-AQA and AQA-7}. 
 Four cases are visualized in Figure \ref{fig5}. 
 The tracking results generated by SiamMask are very stable and accurate.
 The final predicted scores are very close to the GT score since the TSA module is adopted.
 Interestingly, as shown in Figure \ref{fig5}, the VOT tracker can handle various complex situations, such as the disappearance of athletes (\#02-32), drastic changes in scale (\#056) and synchronous diving (\#082).
 These results show that the tracking strategy perfectly meets the requirements of AQA tasks and verify the effectiveness of the TSA module.

\section{Conclusion}
 In this paper, we present a Tube Self-Attention Network (TSA-Net) for action quality assessment, which is able to capture rich spatio-temporal contextual information in human motion. 
 Specifically, a tube self-attention module is introduced to efficiently aggregate contextual information located in ST-Tube generated by SiamMask.
 Sufficient experiments are performed on three datasets: AQA-7, MTL-AQA, and a dataset proposed by us named FR-FS.
 Experimental results demonstrate that TSA-Net can capture long-range contextual information and achieve high performance with less computational cost.
 In the future, an adaptive mechanism of ST-Tube will be explored to avoid the sensitivity of the TSA-Net to the size issue.

\begin{acks}
This work was supported by Shanghai Municipal Science and Technology Major Project 2021SHZDZX0103 and National Natural Science Foundation of China under Grant 82090052.
\end{acks}

\clearpage
\bibliographystyle{ACM-Reference-Format}
\bibliography{main}

\end{document}